\documentclass[11pt,a4paper]{article}
\usepackage[hyperref]{acl2021}
\usepackage{times}
\usepackage{latexsym}

\usepackage{microtype}

\usepackage{tikz}
\usetikzlibrary{positioning}

\usepackage{amsmath}
\usepackage{amssymb}

\usepackage{cleveref}

\usepackage{hhline}
\usepackage{subcaption}
\usepackage{colortbl}

\usepackage{tabularx}

\usepackage{relsize}

\definecolor{OrangeRed}{rgb}{1,0.3,0}
\newcommand*\rot{\rotatebox{0}}

\aclfinalcopy 
\def\aclpaperid{642} 

\title{Factorising Meaning and Form for Intent-Preserving Paraphrasing}

\author{{Tom Hosking \qquad
  Mirella Lapata} \\
  Institute for Language, Cognition and Computation \\
  School of Informatics, University of Edinburgh \\
  10 Crichton Street, Edinburgh EH8 9AB\\
  \texttt{tom.hosking@ed.ac.uk} \quad \texttt{mlap@inf.ed.ac.uk}}

\date{}

\begin{document}
\maketitle
\begin{abstract}
  We propose a method for generating paraphrases of English questions
  that retain the original intent but use a different surface
  form. Our model combines a careful choice of training objective with
  a principled information bottleneck, to induce a latent encoding
  space that disentangles meaning and form. We train an
  encoder-decoder model to reconstruct a question from a
  \textit{paraphrase} with the same meaning and an \textit{exemplar}
  with the same surface form, leading to separated encoding spaces. We
  use a Vector-Quantized Variational Autoencoder to represent the
  surface form as a set of discrete latent variables, allowing us to
  use a classifier to select a different surface form at test
  time. Crucially, our method does not require access to an external
  source of target exemplars.  Extensive experiments and a human
  evaluation show that we are able to generate paraphrases with a
  better tradeoff between semantic preservation and syntactic novelty
  compared to previous methods.


\end{abstract}

\section{Introduction}

A paraphrase of an utterance is ``an alternative surface form in the
same language expressing the same semantic content as the original
form" \cite{madnanidorr}. For questions, a paraphrase should have the
same intent, and should lead to the same answer as the original, as in
the examples in \Cref{tab:intro}. Question paraphrases are of
significant interest, with applications in data augmentation
\cite{iyyer-etal-2018-adversarial}, query rewriting
\cite{dong-etal-2017-learning-paraphrase} and duplicate question
detection \cite{shah-etal-2018-adversarial}, as they allow a system to
better identify the underlying intent of a user query.

Recent approaches to paraphrasing use information bottlenecks with
VAEs \cite{bowman-etal-2016-generating} or pivot languages
\cite{wieting-gimpel-2018-paranmt} to try to extract the semantics of
an input utterance, before projecting back to a (hopefully different)
surface form. However, these methods have little to no control over
the preservation of the input meaning or variation in the output
surface form. Other work has specified the surface form to be
generated
\cite{iyyer-etal-2018-adversarial,chen-etal-2019-controllable,sgcp2020},
but has so far assumed that the set of valid surface forms is known a
priori.


\begin{table}[t]
    \centering
    \small
    \begin{tabular}{cc}
        \hhline{=}
        How is a dialect different from a language? \\
        The differences between language and dialect? \\
        What is the difference between language and dialect? \\

        \hhline{=}
        What is the weight of an average moose? \\
        Average weight of the moose? \\
        How much do moose weigh? \\
        How heavy is a moose? \\

        \hhline{=}
        What country do parrots live in? \\
        In what country do parrots live? \\ 
        Where do parrots naturally live? \\
        What part of the world do parrots live in? \\
        \hhline{=}
        
    \end{tabular}
    \caption{Examples of question paraphrase clusters, drawn from Paralex \cite{fader-etal-2013-paraphrase}. Each member of the cluster has essentially the same semantic \textit{intent}, but a different \textit{surface form}. Each cluster exhibits variation in word choice, syntactic structure and even question type. Our task is to generate these different surface forms, using only a single example as input.} 
    \label{tab:intro}
\end{table}





\begin{figure*}[ht]
    \centering

    \includegraphics[width=0.95\textwidth]{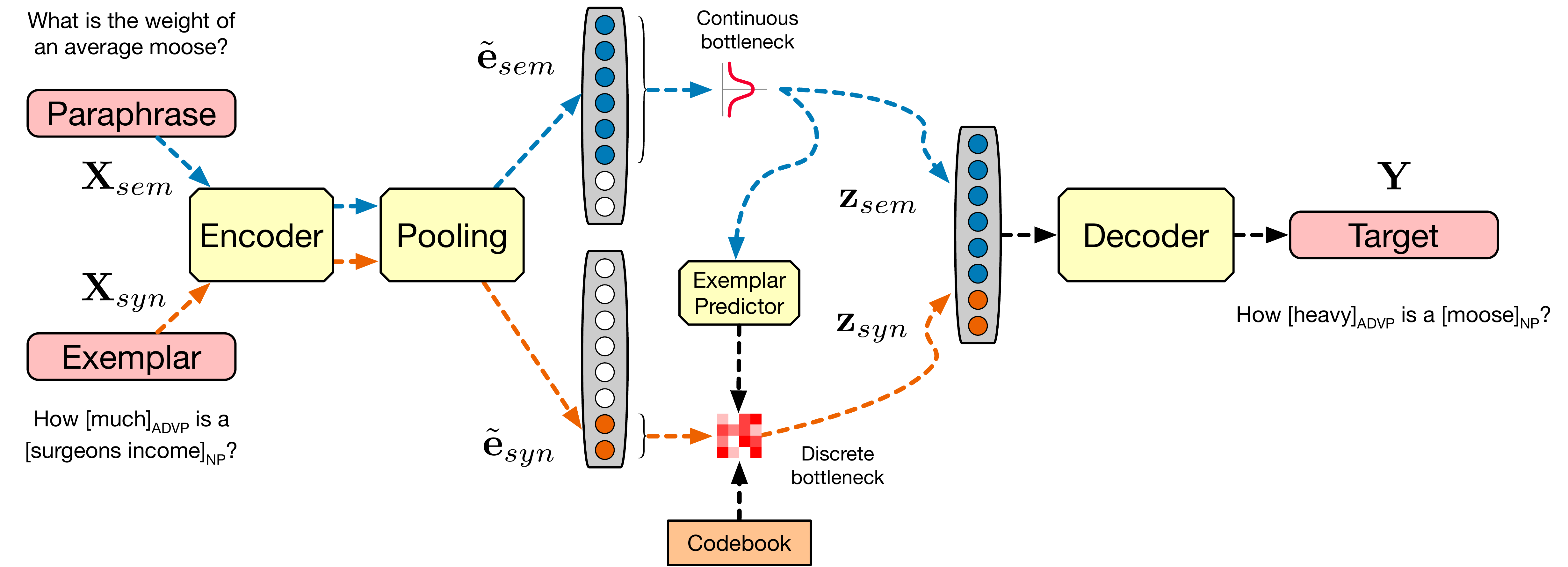}
    \caption{Overview of our approach. The model is trained to
      reconstruct a target question from one input with the same
      \textit{meaning} and another input with the same
      \textit{form}. This induces separate latent encoding spaces for
      meaning and form, allowing us to vary the output form while
      keeping the meaning constant. Using a discretized space for the syntactic encoding makes it tractable to predict valid surface forms at test time.}
    \label{fig:pipeline}
\end{figure*}

In this paper, we propose \textsc{Separator}, a method for generating
paraphrases that exhibit high variation in surface form while still
retaining the original intent. Our key innovations are: (a) to train a model
to reconstruct a target question from an input \textit{paraphrase} with the same
meaning, and an \textit{exemplar} with the same surface form, and (b) to separately encode the form and meaning of
questions as discrete and continuous latent variables respectively, enabling us to modify the output surface form
while preserving the original question intent.   Crucially, unlike
prior work on syntax controlled paraphrasing, we show that we can
generate diverse paraphrases of an input question at test time by
inferring a different discrete syntactic encoding, without needing access to
reference exemplars. 

We limit our work to English questions for three reasons: (a) the concept of a
paraphrase is more clearly defined for questions compared to generic
utterances, as question paraphrases should lead to the same answer;
(b) the space of possible surface forms is smaller for questions,
making the task more achievable, and (c) better dataset
availability. However, our approach does not otherwise make any
assumptions specific to questions.




\section{Problem Formulation}




The task is to learn a mapping from an input question, represented as
a sequence of tokens $\textbf{X}$, to paraphrase(s) $\textbf{Y}$ which
have different \textit{surface form} to~$\textbf{X}$, but convey the
same \textit{intent}.

Our proposed approach, which we call \mbox{\textsc{Separator}}, uses an
encoder-decoder model to transform an input question into a latent
encoding space, and then back to an output paraphrase. We hypothesize
that a principled information bottleneck (\Cref{sec:bottleneck}) and a
careful choice of training scheme (\Cref{sec:training}) lead to an
encoding space that separately represents the intent and surface
form. This separation enables us to paraphrase the input question,
varying the surface form of the output by directly manipulating the
syntactic encoding of the input and keeping the semantic encoding
constant (\Cref{sec:mlp}). We assume access to reference
\textit{paraphrase clusters} during training (e.g.,~\Cref{tab:intro}),
sets of questions with different surface forms that have been collated
as having the same meaning or \textit{intent}.





Our model is a variant of the standard encoder-decoder framework
\cite{Cho2014a}, and consists of: (a) a vanilla Transformer sentence
encoder \cite{Vaswani2017}, that maps an input question $\textbf{X}$
to a multi-head sequence of encodings, $\textbf{e}_{h,t} =
\textsc{Encoder}(\textbf{X})$; (b) a principled choice of information
bottleneck, with a continuous variational path and a discrete vector-quantized
path, that maps the encoding sequence to a pair of latent vectors,
$\textbf{z}_{sem}, \textbf{z}_{syn} =
\textsc{Bottleneck}(\textbf{e}_{h,t})$, represented in more detail in
\Cref{fig:pipeline}; (c) a vanilla Transformer decoder, that attends
over the latent vectors to generate a sequence of output tokens,
$\hat{\textbf{Y}} = \textsc{Decoder}(\textbf{z}_{sem},
\textbf{z}_{syn})$. The separation between $\textbf{z}_{sem}$ and
$\textbf{z}_{syn}$ is induced by our proposed training scheme, shown
in \Cref{fig:pipeline} and described in detail in \Cref{sec:training}.

\subsection{Model Architecture}
\label{sec:bottleneck}

While the encoder and decoder used by the model are standard
Transformer modules, our bottleneck is more complex and we now describe it in more detail.

Let the encoder output be $\{\textbf{e}_{h,1}, \ldots,
\textbf{e}_{h,|\textbf{X}|}\} = \textsc{Encoder}(\textbf{X})$, where
$\textbf{e}_{h,t} \in \mathbb{R}^{D/H_T}$, $h \in {1,
..., H_T}$ with $H_T$
the number of transformer heads, $|\textbf{X}|$ the length of the
input sequence and $D$ the dimension of the transformer. We first pool
this sequence of encodings to a single vector, using the multi-head
pooling described in \newcite{liu-lapata-2019-hierarchical}. For each
head $h$, we calculate a distribution over time indexes $\alpha_{h,t}$
using attention:

\begin{align}
    \alpha_{h,t} &= \frac{\exp{a_{h,t}}}{\sum_{t' \in |\textbf{X}|} \exp{a_{{h,t'}}}}, \\ 
    a_{h,t} &= \textbf{k}_{h}^T  \textbf{e}_{h,t},
\end{align}
with $\textbf{k}_{h} \in \mathbb{R}^{D/H}$ a learned parameter. 

We then take a weighted average of a linear projection of the
encodings, to give pooled output~$\tilde{\textbf{e}}_h$,
\begin{equation}
    \tilde{\textbf{e}}_h = \sum_{t' \in |\textbf{X}|} \alpha_{h,t'} V_h \textbf{e}_{h,t'},
\end{equation}
with $V_{h} \in \mathbb{R}^{D/H \times D/H}$ a learned parameter.

Transformer heads are assigned either to a \textit{semantic} group $H_{sem}$, that
will be trained to encode the intent of the input,
$\tilde{\textbf{e}}_{sem} = [\ldots;\tilde{\textbf{e}}_h; \ldots ], h
\in H_{sem}$, or to a \textit{syntactic} group $H_{syn}$, that will be
trained to represent the surface form $\tilde{\textbf{e}}_{syn} =
[\ldots;\tilde{\textbf{e}}_h; \ldots ], h \in H_{syn}$ (see
\Cref{fig:pipeline}). 

The space of possible question intents is
extremely large and may be reasonably approximated by a continuous
vector space. However, the possible surface forms are discrete and
smaller in number. We therefore use a
Vector-Quantized Variational Autoencoder \citep[VQ-VAE,][]{vqvae} for
the syntactic encoding $\textbf{z}_{syn}$, and model the semantic encoding
$\textbf{z}_{sem}$ as a continuous Gaussian latent variable, as shown in the upper and
lower parts of \Cref{fig:pipeline}, respectively.

\paragraph{Vector Quantization}


Let $q_h$ be discrete latent variables
corresponding to the syntactic quantizer heads, $h \in H_{syn}$.\footnote{The
  number and dimensionality of the quantizer heads need not be the
  same as the number of transformer heads.} Each variable can be one
of $K$ possible latent codes, $q_h \in [0, K]$. The heads use distinct
codebooks, $\textbf{C}_h \in \mathbb{R}^{K \times D/H}$,
which map each discrete code to a continuous embedding
$\textbf{C}_h(q_h) \in \mathbb{R}^{D/H}$. Given sentence
$\textbf{X}$ and its pooled encoding
$\{\tilde{\textbf{e}}_1,...,\tilde{\textbf{e}}_{H}\}$,
we independently quantize the syntactic subset of the heads $h
\in H_{syn}$ to their nearest codes from $\textbf{C}_h$ and
concatenate, giving the syntactic encoding
\begin{equation}
  \textbf{z}_{syn} = [\textbf{C}_1(q_1); \ldots;\textbf{C}_{|H_{syn}|}(q_{|H_{syn}|})].
\end{equation}

The quantizer
module is trained through backpropagation using straight-through
estimation \cite{Bengio2013EstimatingOP}, with an additional loss term
to constrain the embedding space as described in \citet{vqvae},
\begin{equation}
  \mathcal{L}_{cstr} = \lambda \sum_{h \in H_{syn}} \left \lVert   \Big ( \tilde{\textbf{e}}_{h} - \text{sg}(\textbf{C}_h(q_h)) \Big ) \right \rVert_2 ,
\end{equation}
where the stopgradient operator $\text{sg}(\cdot)$ is defined as identity during
forward computation and zero on backpropagation, and $\lambda$ is a weight that controls the strength of the constraint. We follow the soft EM and exponentially moving averages training
approaches described in earlier work
\cite{roy-theory-experiments,angelidis2020extractive}, which we find improve training stability. 

\paragraph{Variational Bottleneck}

For the semantic path, we introduce a learned Gaussian posterior, that
represents the encodings as smooth distributions in space instead of
point estimates \cite{kingma2013autoencoding}.  Formally,
${\phi(\textbf{z}_h} | \textbf{e}_h ) \sim
\mathcal{N}(\boldsymbol{\mu}(\textbf{e}_h),
\boldsymbol{\sigma}(\textbf{e}_h))$, where $\boldsymbol{\mu} ( \cdot
)$ and $\boldsymbol{\sigma} ( \cdot )$ are learned linear
transformations. To avoid vanishingly small variance and to encourage
a smooth distribution, a prior is introduced,
$\textbf{p}(\textbf{z}_h) \sim \mathcal{N}(\textbf{0},
\textbf{1})$. The VAE objective is the standard evidence lower bound
(ELBO), given by
\begin{multline} \label{eq:vaeobjective}
    \text{ELBO} = - \text{KL} [ \phi(\textbf{z}_h | \textbf{e}_h )|| p(\textbf{z}_h )  ] \\
    + \mathbb{E}_{\phi}[ \log p(\textbf{e}_h | \textbf{z}_h ) ].
\end{multline}

We use the usual Gaussian reparameterisation trick, and approximate the expectation in
\Cref{eq:vaeobjective} by sampling from the training set and updating
via backpropagation \cite{kingma2013autoencoding}. The VAE component therefore only adds an
additional KL term to the overall loss,
\begin{equation}
    \mathcal{L}_{KL} = - \text{KL} [\phi(\textbf{z}_h | \textbf{e}_h )|| p(\textbf{z}_h )  ].
\end{equation}

In sum, $\textsc{Bottleneck}(\textbf{e}_{h,t})$ maps a sequence of
token encodings to a pair of vectors $\textbf{z}_{sem},
\textbf{z}_{syn}$, with $\textbf{z}_{sem}$ a continuous latent
Gaussian, and $\textbf{z}_{syn}$ a combination of discrete code
embeddings.

\subsection{Factorised Reconstruction Objective}
\label{sec:training}

We now describe the training scheme that causes the model to learn
separate encodings for meaning and form: $\textbf{z}_{sem}$ should
encode only the intent of the input, while $\textbf{z}_{syn}$ should
capture any information about the surface form of the input. Although
we refer to $\textbf{z}_{syn}$ as the \textit{syntactic encoding}, it
will not necessarily correspond to any specific syntactic
formalism. We also acknowledge that meaning and form are not
completely independent of each other; arbitrarily changing the form of
an utterance is likely to change its meaning. However, it is possible
for the same intent to have multiple phrasings , and it is this `local
independence' that we intend to capture.

We create triples $\{\textbf{X}_{sem}, \textbf{X}_{syn},
\textbf{Y}\}$, where $\textbf{X}_{sem}$ has the same meaning but
different form to $\textbf{Y}$ (i.e.,~it is a paraphrase, as in
\Cref{tab:intro}) and $\textbf{X}_{syn}$ is a question with the same
form but different meaning (i.e.,~it shares the same syntactic
\textit{template} as $\textbf{Y}$), which we refer to as an
\textit{exemplar}. We describe the method for retrieving these
exemplars in \Cref{sec:exemplars}. The model is then trained to
generate a target paraphrase $\textbf{Y}$ from the semantic encoding
$\textbf{z}_{sem}$ of the input paraphrase $\textbf{X}_{sem}$, and
from the syntactic encoding $\textbf{z}_{syn}$ of the exemplar
$\textbf{X}_{syn}$, as demonstrated in \Cref{fig:pipeline}.

Recalling the additional losses from the variational and quantized
bottlenecks, the final combined training objective is given by
\begin{equation}
    \mathcal{L} = \mathcal{L}_{\textbf{Y}} + \mathcal{L}_{cstr} + \mathcal{L}_{KL},
\end{equation}
where $\mathcal{L}_{\textbf{Y}}(\textbf{X}_{sem}, \textbf{X}_{syn})$ is the cross-entropy loss of teacher-forcing the decoder to generate $\textbf{Y}$ from $\textbf{z}_{sem}(\textbf{X}_{sem})$ and  $\textbf{z}_{syn}(\textbf{X}_{syn})$.






\begin{table}[t!]
    \small
    \centering
    \begin{tabular}{@{}c@{~}|@{~}c@{}}
    
    \hhline{==}
        \textit{Input} & How heavy is a moose? \\
        \textit{Chunker output} & How [heavy]\textsubscript{ADVP} is a [moose]\textsubscript{NP} ? \\
        \textit{Template} & How ADVP is a NP ? \\
        \textit{Exemplar} & How much is a surgeon's income? \\
    \hhline{==}
        \textit{Input} & What country do parrots live in \\
        \textit{Chunker output} & What [country]\textsubscript{NP} do  [parrots]\textsubscript{NP} [live]\textsubscript{VP} in ? \\
        \textit{Template} & What NP do NP VP in ? \\
        \textit{Exemplar} & What religion do Portuguese believe in? \\
     \hhline{==}
    \end{tabular}
\vspace*{-.1cm}
    \caption{Examples of the exemplar retrieval process for training. The input is tagged by a chunker, ignoring stopwords. An exemplar with the same template is then retrieved from a different paraphrase cluster.}
    \label{tab:templateexample}
\end{table}

\subsection{Exemplars}
\label{sec:exemplars}

It is important to note that not all surface forms are valid or
\textit{licensed} for all question intents. As shown in
\Cref{fig:pipeline}, our approach requires exemplars during training
to induce the separation between latent spaces.  We also need to
specify the desired surface form at \textit{test time}, either by
supplying an exemplar as input or by directly predicting the latent
codes. The output should have a different surface form to the input
but remain fluent.

\paragraph{Exemplar Construction}

During training, we retrieve exemplars $\textbf{X}_{syn}$ from the
training data following a process which first identifies the
underlying syntax of $\textbf{Y}$, and finds a question with the same
syntactic structure but a different, arbitrary meaning. We use a
shallow approximation of syntax, to ensure the availability of
equivalent exemplars in the training data. An example of the exemplar
retrieval process is shown in \Cref{tab:templateexample}; we first
apply a chunker \citep[FlairNLP, ][]{akbik-etal-2018-contextual} to
$\textbf{Y}$, then extract the chunk label for each tagged span,
ignoring stopwords. This gives us the \textit{template} that
$\textbf{Y}$ follows. We then select a question at random from the
training data with the same template to give $\textbf{X}_{syn}$. If no
other questions in the dataset use this template, we create an exemplar by replacing each chunk with a random sample of the same type.


We experimented with a range of approaches to determining question
templates, including using part-of-speech tags and (truncated)
constituency parses. We found that using chunks and preserving
stopwords gave a reasonable level of granularity while still combining
questions with a similar form. The templates (and corresponding
exemplars) need to be granular enough that the model is forced to use
them, but abstract enough that the task is not impossible to learn.

\paragraph{Prediction at Test Time}
\label{sec:mlp}


In general, we do not assume access to reference exemplars at test
time and yet the decoder must generate a paraphrase from semantic
\emph{and} syntactic encodings.  Since our latent codes are separated,
we can \emph{directly} predict the syntactic encoding, without needing
to retrieve or generate an exemplar. Furthermore, by using a discrete
representation for the syntactic space, we reduce this prediction
problem to a simple classification task. Formally, for an input
question $\textbf{X}$, we learn a distribution over licensed discrete
codes $q_h, h \in \tilde{H}_{syn}$. We assume that the heads are
independent, so that $p(q_1,\ldots,q_{\tilde{H}_{syn}}) = \prod_{i}
p(q_i)$. We use a small fully connected network with the semantic and
syntactic encodings of $\textbf{X}$ as inputs, giving $p(q_h |
\textbf{X}) = \textsc{MLP}(\textbf{z}_{sem}(\textbf{X}),
\textbf{z}_{syn}(\textbf{X}))$.

The network is trained to maximize the likelihood of all other
syntactic codes licensed by each input. We calculate the discrete
syntactic codes for each question in a paraphrase cluster, and
minimize the cross-entropy loss of the network with respect to these
codes. At test time, we set $q_h = \text{argmax}_{q'_h} [p(q'_h |
\textbf{X}_{test})]$.




\section{Experimental Setup}

\paragraph{Datasets}

We evaluate our approach on two datasets: Paralex
\cite{fader-etal-2013-paraphrase}, a dataset of question paraphrase
clusters scraped from WikiAnswers; and Quora Question Pairs
(QQP)\footnote{\mbox{\url{https://www.kaggle.com/c/quora-question-pairs}}}
sourced from the community question answering forum Quora.  We
observed that a significant fraction of the questions in Paralex
included typos or were ungrammatical. We therefore filter out any
questions marked as non-English by a language detection script
\cite{lui-baldwin-2012-langid}, then pass the questions through a
simple spellchecker. While this destructively edited some named
entities in the questions, it did so in a consistent way across the
whole dataset. There is no canonical split for Paralex, so we group
the questions into clusters of paraphrases, and split these clusters
into train/dev/test partitions with weighting 80/10/10. Similarly, QQP
does not have a public test set. We therefore partitioned the
clusters in the validation set randomly in two, to give us our
dev/test splits. Summary statistics of the resulting datasets are
given in \Cref{app:dataset}.  All scores reported are on our test
split.

\paragraph{Model Configuration}

Following previous work
\cite{kaiser-latent-variables,angelidis2020extractive}, our quantizer uses
multiple heads ($H=4$) with distinct codebooks to represent the syntactic encoding as 4~discrete
categorical variables $q_h$, with $\textbf{z}_{syn}$ given by the
concatenation of their codebook embeddings $\textbf{C}_h(q_h)$. We use a relatively small
codebook size of $K=256$, relying on the combinatoric power of the
multiple heads to maintain the expressivity of the model. We argue
that, assuming each head learns to capture a particular property of a
template (see \Cref{sec:heads}), the number of variations in
\textit{each property} is small, and it is only through combination
that the space of possible templates becomes large.




We include a detailed list of hyperparameters in \Cref{app:hyperparams}. Our code is available at \url{http://github.com/tomhosking/separator}.

\paragraph{Comparison Systems}

We compare \textsc{Separator} against several related systems.  These
include a model which reconstructs $\textbf{Y}$ only from
$\textbf{X}_{sem}$, with no signal for the desired form of the
output. In other words, we derive both $\textbf{z}_{sem}$ and
$\textbf{z}_{syn}$ from $\textbf{X}_{sem}$, and \textit{no separation}
between meaning and form is learned. This model uses a continuous
Gaussian latent variable for both $\textbf{z}_{syn}$ and
$\textbf{z}_{sem}$, but is otherwise equivalent in architecture to
\textsc{Separator}. We refer to this as the \textit{VAE} baseline. We
also experiment with a vanilla autoencoder or \textit{AE} baseline by
removing the variational component, such that $\textbf{z}_{sem},
\textbf{z}_{syn} = \tilde{\textbf{e}}_{sem},
\tilde{\textbf{e}}_{syn}$.

We include our own implementation of the \mbox{VQ-VAE} model described
in \citet{roy-grangier-2019-unsupervised}. They use a quantized
bottleneck for \textit{both} $\textbf{z}_{sem}$ and
$\textbf{z}_{syn}$, with a large codebook $K = 64,000$, $H = 8$ heads and a residual
connection within the quantizer. For QQP, containing only 55,611
training clusters, the configuration in
\citet{roy-grangier-2019-unsupervised} leaves the model
overparameterized and training did not converge; we instead report
results for $K = 1,000$.

ParaNMT \cite{wieting-gimpel-2018-paranmt} translates input sentences
into a pivot language (Czech), then back into English. Although this
system was trained on high volumes of data (including Common Crawl),
the training data contains relatively few questions, and we would not
expect it to perform well in the domain under consideration. `Diverse
Paraphraser using Submodularity' (DiPS;
\citealt{kumar-etal-2019-submodular}) uses submodular optimisation to
increase the diversity of samples from a standard encode-decoder
model. Latent bag-of-words (BoW; \citealt{latentbow}) uses an
encoder-decoder model with a discrete bag-of-words as the latent
encoding. SOW/REAP \cite{goyal_neural_2020} uses a two stage approach, deriving a set of feasible syntactic rearrangements that is used to guide a second encoder-decoder model. We additionally implement a simple \mbox{tf-idf} baseline
\cite{tfidf}, retrieving the question from the training set with the
highest similarity to the input. Finally, we include a basic copy
baseline as a lower bound, that simply uses the input question as the output.



\begin{table}[t!]
    \begin{subtable}[]{0.49\textwidth}
        \centering
        \begin{tabular}{c||cc}
        & \multicolumn{2}{c}{\textit{Cluster type}} \\
       \textbf{Encoding} & \textbf{Paraphrase} & \textbf{Template} \\
        \hhline{=#==}
        $\textbf{z}_{sem}$ & \cellcolor{OrangeRed!94}\color{white}0.943 & \cellcolor{OrangeRed!9}0.096 \\
        $\textbf{z}_{syn}$ & \cellcolor{OrangeRed!95}\color{white}0.952 & \cellcolor{OrangeRed!9}0.092 \\
        \hline
        $\textbf{z}$ & \cellcolor{OrangeRed!96}\color{white}\textbf{0.960} & \cellcolor{OrangeRed!9}\textbf{0.096} \\
        \end{tabular}
        \caption{VAE Baseline}
        \label{tab:separation_baseline}
    \end{subtable}
    
\vspace{.1cm}
  \hfill
    \begin{subtable}[]{0.49\textwidth}
    
        \centering
        \begin{tabular}{c||cc}
        & \multicolumn{2}{c}{\textit{Cluster type}} \\
       \textbf{Encoding} & \textbf{Paraphrase} & \textbf{Template} \\
        \hhline{=#==}
        $\textbf{z}_{sem}$ & \cellcolor{OrangeRed!94}\color{white}\textbf{0.944} & \cellcolor{OrangeRed!5}0.053 \\
        $\textbf{z}_{syn}$ & \cellcolor{OrangeRed!6}0.065 & \cellcolor{OrangeRed!86}\color{white}\textbf{0.866} \\
        \hline
        $\textbf{z}$ & \cellcolor{OrangeRed!31}0.307 & \cellcolor{OrangeRed!85}\color{white}0.849 \\
        \end{tabular}
        \caption{\textsc{Separator}}
        \label{tab:separation_ours}
    \end{subtable}
    \caption{Retrieval accuracies for each encoding for each cluster type. The VAE baseline is trained only on paraphrase pairs and receives no signal for the desired form of the output. \textsc{Separator} is able to learn separate encodings for meaning and form, with negligible loss in semantic encoding performance.}
    \label{tab:separation}
\end{table}




\begin{table*}[ht!]
    \centering
    \begin{tabular}{l|rrr|rrr}
     & \multicolumn{3}{c|}{\textbf{Paralex}} & \multicolumn{3}{c}{\textbf{QQP}} \\
    \textbf{Model} & \rot{BLEU} $\uparrow$ & \rot{Self-BLEU} $\downarrow$  & \rot{\textbf{iBLEU}} $\uparrow$  & \rot{BLEU} $\uparrow$  & \rot{Self-BLEU} $\downarrow$ & \rot{\textbf{iBLEU}} $\uparrow$ \\
\hline \hline
    Copy  & 37.10 & 100.00 & $-$4.03 & 32.61 & 100.00 & $-$7.17 \\ 
    VAE  & 40.26 & 66.12 & 8.35 & 19.36 & 35.29 & 2.96 \\ 
    AE  & 40.10 & 75.71 & 5.36 & 19.90 & 39.81 & 1.99 \\ 
    \mbox{tf-idf}  & 25.08 & 25.25 & 9.98 & 22.73 & 61.81 & $-$2.63 \\ 
    VQ-VAE  & 40.26 & 65.71 & 8.47 & 16.19 & 26.15 & 3.43 \\ 
    \hline\hline
    ParaNMT & 20.42 & 39.90 & 2.32 & 24.24 & 56.42 & 0.04 \\ 
    DiPS  & 24.90 & 29.58 & 8.56 & 18.47 & 32.45 & 3.19 \\ 
    SOW/REAP & 33.09 & 37.07 & 12.04 & 12.64 & 24.19 & 1.59 \\ 
    LBoW & 34.96 & 35.86 & 13.71 & 16.17 & 29.00 & 2.62 \\ 
    \textsc{Separator} & 36.36 & 35.37 & \textbf{14.84} & 14.70 & 14.84 & \textbf{5.84} \\
    \hline\hline 
    \textsc{Oracle} & 53.37 & 24.55 & 29.99 & 24.50 & 16.04 &
    12.34 \\
    \hline \hline
    \end{tabular}
\vspace*{-.2cm}
    \caption{Generation results, without access to oracle
      exemplars. Our approach achieves the highest iBLEU scores, indicating the best   tradeoff between output diversity and fidelity to the reference
      paraphrases.}  
    \label{tab:ibleu}
\end{table*}


\section{Results}

Our experiments were designed to answer three questions: (a)~Does
\textsc{Separator} effectively factorize meaning and form? (b)~Does \textsc{Separator}
manage to generate diverse paraphrases (while preserving the intent of
the input)? (c)~What does the underlying quantized space encode
(i.e.,~can we identify any meaningful syntactic properties)? We
address each of these questions in the following sections.

\subsection{Verification of Separation}

Inspired by \citet{chen-etal-2019-multi} we use a \textit{semantic textual similarity} task and a \textit{template detection} task to confirm that \textsc{Separator} does indeed lead to encodings $\{\textbf{z}_{sem}, \textbf{z}_{syn}\}$ in latent spaces that represent different types of information. 

Using the test set, we construct clusters of questions that share the
same meaning $\mathcal{C}_{sem}$, and clusters that share the same
template $\mathcal{C}_{syn}$. For each cluster $C_q \in
\{\mathcal{C}_{sem}, \mathcal{C}_{syn}\}$, we extract one question at
random $\textbf{X}_q \in C_q$, compute its encodings
$\{\textbf{z}_{sem}, \textbf{z}_{syn},
\textbf{z}\}$\footnote{$\textbf{z}$ refers to the combined encoding,
i.e.,~$[\textbf{z}_{sem}; \textbf{z}_{syn}]$.}, and its cosine
similarity to the encodings of all other questions in the test set. We
take the question with maximum similarity to the query $\textbf{X}_r,
r = \text{argmax}_{r'}( \textbf{z}_q . \textbf{z}_{r'})$, and compare
the cluster that it belongs to, $C_r$, to the query cluster $I(C_q =
C_r)$, giving a \textit{retrieval} accuracy score for each encoding
type and each clustering type. For the VAE, we set
$\{\textbf{z}_{sem}, \textbf{z}_{syn}\}$ to be the same heads of
$\textbf{z}$ as the separated model.


\Cref{tab:separation} shows that our approach yields encodings that successfully factorise meaning and form, with negligible performance loss compared to the VAE baseline; paraphrase retrieval performance using $\textbf{z}_{sem}$ for the separated model is comparable to using $\textbf{z}$ for the VAE.






\subsection{Paraphrase Generation}

\paragraph{Automatic Evaluation} 
While we have shown that our approach leads to disentangled
representations, we are ultimately interested in generating diverse
paraphrases for \textit{unseen data}. That is, given some input
question, we want to generate an output question with the same meaning
but different form.

We use iBLEU \cite{ibleu} as our primary metric, a variant of BLEU
\cite{papineni-etal-2002-bleu,post-2018-call} that is penalized by the
similarity between the output and the \textit{input}, 
\begin{align}
\begin{split}
    \textrm{iBLEU} = \alpha \textrm{BLEU}(output, references) \\- (1-\alpha) \textrm{BLEU}(output, input),
\end{split}
\end{align}
where $\alpha = 0.7$ is a constant that weights the tradeoff between
fidelity to the references and variation from the input. We also
report the usual $\textrm{BLEU}(output, references)$ as well as
\mbox{Self-{BLEU}}$(output, input)$. The latter allows us to examine
whether the models are making trivial changes to the input. The Paralex
test set contains 5.6 references on average per cluster, while QQP
contains only 1.3. This leads to lower BLEU scores for QQP in general,
since the models are evaluated on whether they generated the
specific paraphrase(s) present in the dataset.





\Cref{tab:ibleu} shows that the Copy, VAE and AE models display
relatively high BLEU scores, but achieve this by `parroting' the
input; they are good at reconstructing the input, but introduce little
variation in surface form, reflected in the high Self-BLEU
scores. This highlights the importance of considering similarity to
both the references \textit{and} to the input.  The \mbox{tf-idf}
baseline performs surprisingly well on Paralex; the large dataset size makes it more likely that a paraphrase cluster with a similar meaning to the query exists in the training set.  

The other comparison
systems (in the second block in \Cref{tab:ibleu}) achieve lower Self-BLEU
scores, indicating a higher degree of variation introduced, but this
comes at the cost of much lower scores with respect to the
references. \textsc{Separator} achieves the highest iBLEU scores,
indicating the best balance between fidelity to the references and
novelty compared to the input. We give some example output in
\Cref{tab:examples}; while the other systems mostly introduce lexical
variation, \textsc{Separator} is able to produce output with markedly different syntactic structure to the input, and can even
change the question type while successfully preserving the original intent.





\begin{table}[t]
    \small
    \centering
    \begin{tabular}{@{}r@{~}|@{~}p{6cm}@{}}
    \hhline{==} 
    \textit{Input} & What is the most known singer? \\
    \hline
    {VAE} & What is the most known singer? \\
    {DiPS} & What was the most known famous singer? \\
    {SOW/REAP} & What is the most famous singer? \\
    {Latent BoW} & What is the most famous singer? \\
    \textsc{Separator} & Who is the most famous singer in America? \\
    \hhline{==} 
    \textit{Input} & What is the income for a soccer player? \\
    \hline
    {VAE} &  What is the salary for a soccer player? \\
    {DiPS} &  What is the median income in soccer? \\
    {SOW/REAP} & What is US cer? \\
    {Latent BoW} &  What is the salary of a soccer [UNK]? \\
    \textsc{Separator} &  How much is a soccer players' salary? \\
    \hhline{==} 
    \textit{Input} & What has been the economic impact from Brexit referendum so far? \\
    \hline
    {VAE} &  What has been the economic impact of Brexit referendum so far? \\
    {DiPS} &  What will be a impact of Brexit referendum? \\
    {SOW/REAP} & How do I spend my virginity? \\
    {Latent BoW} &  How did Brexit referendum impact the Brexit referendum? \\
    \textsc{Separator} &  How much will the Brexit referendum cost? \\
    \hhline{==} 
    \textit{Input} & What are the basics I should know before learning Hadoop? \\
    \hline
    {VAE} &  What are the basics should I know before learning Hadoop? \\
    {DiPS} &  How do I know before I want to learn Hadoop? \\
    {SOW/REAP} & How can I know before learning Hadoop? \\
    {Latent BoW} &  What are the basics of learning Hadoop? \\
    \textsc{Separator} &  How much should I know before learning Hadoop? \\
    \hhline{==} 
    \end{tabular}
\vspace*{-.1cm}
    \caption{Examples of output generated by various approaches for a
      given input, from Paralex and QQP. \textsc{Separator} is able to
      generate questions with a different syntactic form to the input.}
    \label{tab:examples} 
\end{table}








The last row in \Cref{tab:ibleu} (\textsc{Oracle}) reports results when our model is
given a valid exemplar to use directly for generation, thus bypassing
the code prediction problem.  For each paraphrase cluster, we select
one question at random to use as input, and select another to use as
the target. We retrieve a question from the training set with the
same template as the target to use as an \textit{oracle
  exemplar}. This represents an upper bound on our model's performance.
While \textsc{Separator} outperforms existing methods, our approach to
predicting syntactic codes (using a shallow fully-connected network) is
relatively simple. \textsc{Separator} using oracle exemplars achieves
by far the highest scores in \Cref{tab:ibleu}, demonstrating the
potential expressivity of our approach when exemplars are guaranteed
to be valid. A more powerful code prediction model could close the gap
to this upper bound, as well as enabling the generation of multiple
diverse paraphrases for a single input question. However, we leave
this to future work.


\paragraph{Human Evaluation}


In addition to automatic evaluation we elicited judgements from
crowdworkers on Amazon Mechanical Turk. Specifically, they were shown
a question and two paraphrases thereof (corresponding to different
systems) and asked to select which one was preferred along three
dimensions: the \textit{dissimilarity} of the paraphrase compared to
the original question, how well the paraphrase reflected the
\textit{meaning} of the original, and the \textit{fluency} of the
paraphrase (see \Cref{app:humeval}). We evaluated a total of 200
questions sampled equally from both Paralex and QQP, and collected 3
ratings for each sample. We assigned each system a score of $+1$ when
it was selected, $-1$ when the other system was selected, and took the
mean over all samples. Negative scores indicate that a system was
selected less often than an alternative. We chose the four best
performing models according to \Cref{tab:ibleu} for our evaluation:
\textsc{Separator}, DiPS \cite{kumar-etal-2019-submodular}, Latent BoW
\cite{latentbow} and VAE.


\begin{figure}
    \centering
    \includegraphics[width=0.49\textwidth]{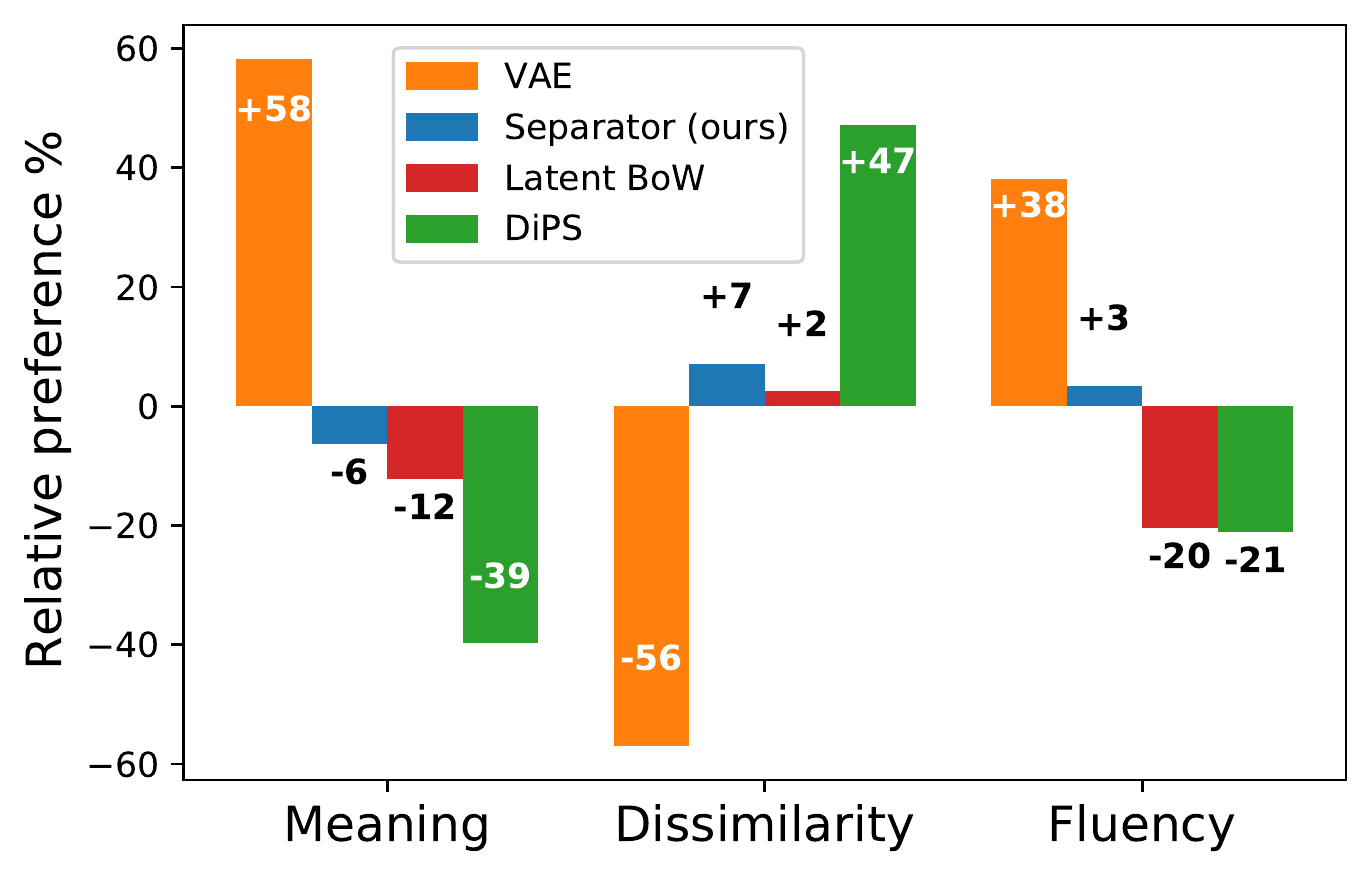}
    \caption{Results of our human evaluation. Although the VAE
      baseline is the best at preserving question meaning, it is the
      worst at introducing variation to the output. \textsc{Separator}
      offers the best balance between dissimilarity and meaning
      preservation, and is more fluent than both DiPS and Latent BoW.} 
    \label{fig:humaneval}
\end{figure}

\Cref{fig:humaneval} shows that although the VAE baseline is the best
at preserving question meaning, it is also the worst at introducing
variation to the output. \textsc{Separator} introduces more variation
than the other systems evaluated and better preserves the original
question intent, as well as generating significantly more fluent
output (using a one-way ANOVA with post-hoc Tukey HSD test,
\mbox{p$<$0.05}).

\subsection{Analysis}


\label{sec:heads}

\begin{figure}[t]
    \centering
    \includegraphics[width=0.4\textwidth]{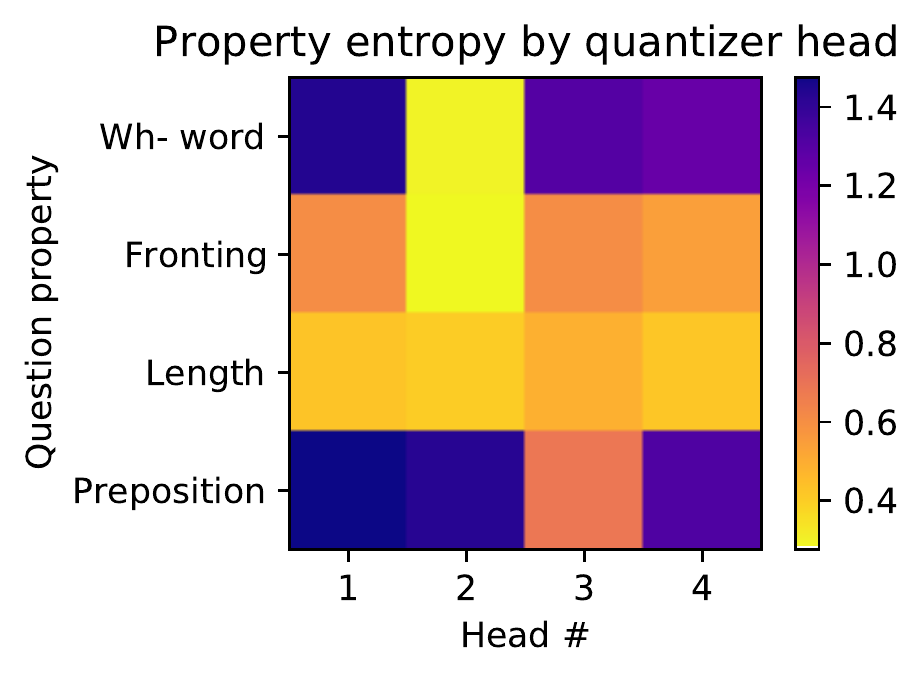}
\vspace*{-.3cm}
    \caption{Predictive entropy by head for various question properties - lower entropy indicates higher predictive power. }
    \label{fig:headentropy}
\end{figure}

When predicting latent codes at test time, we assume that the code for
each head may be predicted independently of the others, as working
with the full joint distribution would be intractable. We now examine
this assumption as well as whether different encodings represent
distinct syntactic properties. Following
\citet{angelidis2020extractive}, we compute the probability of
a question property ${f_1, f_2, \ldots}$ taking a particular value $a$, conditioned by head~$h$ and
quantized code~$k_h$ as
\begin{equation}
\hspace*{-.32cm}
P(f_i | h, k_h)\hspace*{-.3ex}=\hspace*{-.3ex}\frac{\hspace*{-.2ex}\sum\limits_{x \in \mathcal{X}}\hspace*{-.3ex}I(q_h(x)\hspace*{-.3ex}=\hspace*{-.3ex}k_h)  I(f_i(x)\hspace*{-.3ex}=\hspace*{-.3ex}a)}{\sum\limits_{x \in \mathcal{X}} I(q_h(x)\hspace*{-.3ex}=\hspace*{-.3ex}k_h)},\hspace*{-.25cm}
\end{equation}
where $I(\cdot)$ is the indicator function, and examples of values~$a$ are shown in
\Cref{fig:headentropy}. We then calculate the mean entropy of
these distributions, to determine how property-specific each head is:
\begin{equation}
    \hspace*{-.25cm}\mathcal{H}_h = \frac{1}{K} \sum_{k_h}  \sum_a P(a| h, k_h) \log P(a| h, k_h). \hspace*{-.15cm}
\end{equation}

Heads with lower entropies are more predictive of a property,
indicating specialisation and therefore
independence. \Cref{fig:headentropy} shows our analysis for four
syntactic properties: head \#2 has learned to control the high level
output structure, including the question type or \textit{wh- word},
and whether the question word appears at the beginning or end of the
question. Head \#3 controls which type of prepositional phrase is
used. The length of the output is not determined by any one head,
implying that it results from other properties of the surface form.
Future work could leverage this disentanglement to improve the exemplar prediction model, and could lead to more fine-grained control over the generated output form.

In summary, we find that \textsc{Separator} successfully learns separate encodings for meaning and form. \textsc{Separator} is able to generate question paraphrases with a better balance of diversity and intent preservation compared to prior work. Although we are able to identify some high-level properties encoded by each of the syntactic latent variables, further work is needed to learn interpretable syntactic encodings.

\section{Related Work}

\paragraph{Paraphrasing}

Prior work on generating paraphrases has looked at extracting
sentences with similar meaning from large corpora
\cite{barzilay-mckeown-2001-extracting,bannard-callison-burch-2005-paraphrasing,ganitkevitch-etal-2013-ppdb},
or identifying paraphrases from sources that are weakly aligned
\cite{dolan-etal-2004-unsupervised,coster-kauchak-2011-simple}.

More recently, neural approaches to paraphrasing have shown
promise. Several models have used an information bottleneck to try to
encode the semantics of the input, including VAEs
\cite{bowman-etal-2016-generating}, VQ-VAEs
\cite{vqvae,roy-grangier-2019-unsupervised}, and a latent bag-of-words
model \cite{latentbow}. Other work has relied on the strength of
neural machine translation models, translating an input into a
\textit{pivot} language and then back into English
\cite{mallinson-etal-2017-paraphrasing,wieting-gimpel-2018-paranmt,parabank}.

\citet{kumar-etal-2019-submodular} use submodular function maximisation to improve the diversity of paraphrases generated by an encoder-decoder model. \citet{dong-etal-2017-learning-paraphrase} use an automatic paraphrasing system to rewrite inputs to a question answering system at inference time, reducing the sensitivity of the system to the specific phrasing of a query.


\paragraph{Syntactic Templates}

The idea of generating paraphrases by controlling the structure of the output has seen recent interest, but most work so far has assumed access to a template oracle. \citet{iyyer-etal-2018-adversarial} use linearized parse trees as a template, then sample paraphrases by using multiple templates and reranking the output. \citet{chen-etal-2019-controllable} use a multi task objective to train a model to generate output that follows an input template. Their approach is limited by their use of automatically generated paraphrases for training, and their reliance on the availability of oracle templates. \citet{bao-etal-2019-generating} use a discriminator to separate spaces, but rely on noising the latent space to induce variation in the output form. Their results show good fidelity to the references, but low variation compared to the input. \citet{goyal_neural_2020} use the artifically generated dataset ParaNMT-50m \cite{wieting-gimpel-2018-paranmt} for their training and evaluation, which displays low output variation according to our results. \citet{sgcp2020} show strong performance using full parse trees as templates, but focus on generating output with the correct parse and do not consider the problem of template prediction.

\citet{synpg} independently and concurrently propose training a model with a similar `split training' approach to ours, but using constituency parses instead of exemplars, and a `bag-of-words' instead of reference paraphrases. Their approach has the advantage of not requiring paraphrase clusters during training, but they do not attempt to solve the problem of template prediction and rely on the availability of oracle target templates.

\citet{russin-etal-2020-compositional} modify the architecture of an
encoder-decoder model, introducing an inductive bias to encode the
structure of inputs separately from the lexical items to improve
compositional generalisation on an artificial semantic parsing task. \citet{chen-etal-2019-multi} use a multi-task setup to generate separated encodings, but do not experiment with generation tasks. \citet{shu-etal-2019-generating} learn discrete latent codes to introduce variation to the output of a machine translation system.




\section{Conclusion}

We present \textsc{Separator}, a method for generating paraphrases
that balances high variation in surface form with strong intent
preservation. Our approach consists of: (a) a training scheme that
causes an encoder-decoder model to learn separated latent encodings,
(b) a vector-quantized bottleneck that results in discrete variables
for the syntactic encoding, and (c) a simple model to predict
different yet valid surface forms for the output. Extensive
experiments and a human evaluation show that our approach leads to
separated encoding spaces with negligible loss of expressivity, and is
able to generate paraphrases with a better balance of variation and
semantic fidelity than prior methods.

In future, we would like to investigate the properties of the
syntactic encoding space, and improve on the code prediction model. It
would also be interesting to reduce the levels of supervision required
to train the model, and induce the separation without an external
syntactic  model or reference paraphrases. 

\section*{Acknowledgements}

We thank our anonymous reviewers for their feedback. We are grateful
to Stefanos Angelidis for many valuable discussions, and Hao Tang for
their comments on the paper. This work was supported in part by the
UKRI Centre for Doctoral Training in Natural Language Processing,
funded by the UKRI (grant EP/S022481/1) and the University of
Edinburgh. Lapata acknowledges the support of the European Research
Council (award number 681760, ``Translating Multiple Modalities into
Text'').

\bibliography{anthology,acl2020}
\bibliographystyle{acl_natbib}

\clearpage
\appendix

\section{Hyperparameters}
\label{app:hyperparams}

Hyperparameters were selected by manual tuning, based on a combination
of: (a) validation encoding separation, (b) validation BLEU scores
using oracle exemplars, and (c) validation iBLEU scores using
predicted syntactic codes.

\begin{table}[ht]
    \centering
\small
    \begin{tabular}{l|p{2cm}}
    Embedding dimension $D$ & 768 \\
    Encoder layers & 5 \\
    Decoder layers & 5 \\
    Feedforward dimension & 2048 \\
    Transformer heads & 8 \\
    Semantic/syntactic heads $H_{sem},H_{syn}$ & 6/2 \\
    Quantizer heads $\tilde{H}_{syn}$ & 4 \\
    Quantizer codebook size $K$& 256 \\
    Optimizer & Adam \cite{adam} \\
    Learning rate & 0.005 \\
    Batch size & 64 \\
    Token dropout & 0.2 \cite{tokendropout} \\
    Decoder & Beam search \\
    Beam width & 4 \\
    Commitment weight $\lambda$ & 0.25 \\
    \hline
    Code classifier & \\
    Num. hidden layers & 2 \\
    Hidden layer size & 2712 \\
    
    \end{tabular}
    \caption{Hyperparameter values used for our experiments.}
    \label{tab:hyperparams}
\end{table}

\section{Dataset Statistics}
\label{app:dataset}

Summary statistics for our partitions of Paralex and QQP are shown in \Cref{tab:dataset}. Questions in QQP were 9.7 tokens long on average, compared to 8.2 for Paralex.

\begin{figure}[ht]
    \centering
    \includegraphics[width=0.45\textwidth]{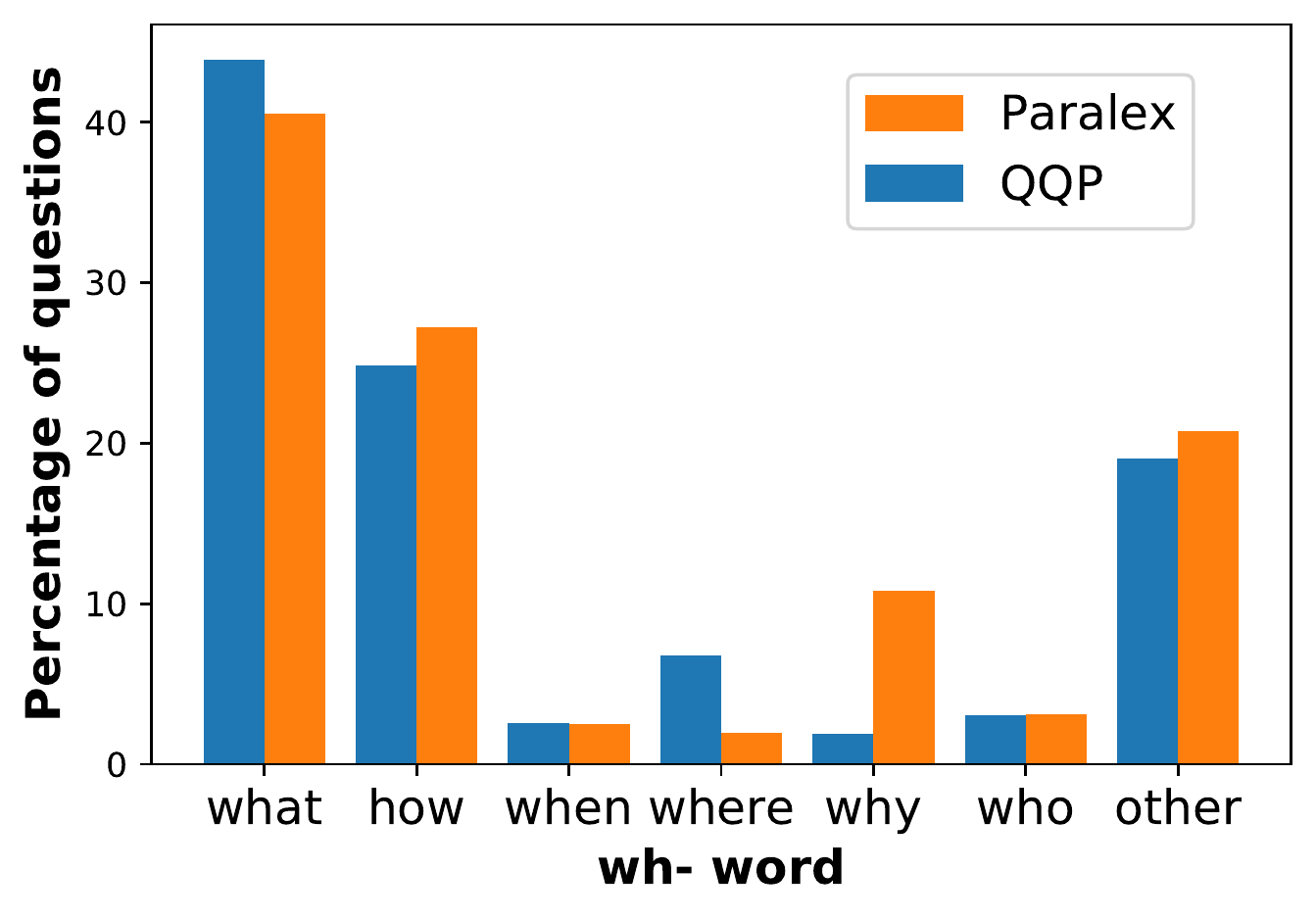}
\vspace*{-.3cm}
    \caption{Distribution of wh- words for the datasets used in our experiments. QQP contains a much higher percentage of \textit{why} questions. }
    \label{fig:whwords}
\end{figure}

We also show the distribution of different question types in \Cref{fig:whwords}; QQP contains a higher percentage of \textit{why} questions, and we found that the questions tend to be more subjective compared to the predominantly factual questions in Paralex.

\begin{table}[ht]
    \centering
\small
    
    \begin{tabular}{l|rr|rr}
        & \multicolumn{2}{c}{\textbf{Paralex}} & \multicolumn{2}{c}{\textbf{QQP}} \\
        & Clusters & Questions & Clusters & Questions \\
        \hline
        \textit{Train} & 222,223 & 1,450,759 & 55,611 & 138,965  \\
        \textit{Dev} & 27,778 & 183,273 & 5,255 & 12,554 \\
        \textit{Test}  & 27,778 & 182,818 & 5,255 & 12,225
    \end{tabular}
    
    \caption{Summary statistics for our cleaned version of \cite{fader-etal-2013-paraphrase}, and our partitioning of QQP.}
    \label{tab:dataset}
\end{table}

\section{Human Evaluation}
\label{app:humeval}

Annotators were asked to rate the outputs according to the following criteria:

\begin{itemize}
    \item Which system output is the most fluent and grammatical?
    \item To what extent is the meaning expressed in the original
      question preserved in the rewritten version, with no additional
      information added? Which of the questions generated by a system
      is likely to have the same answer as the original?
    \item Does the rewritten version use different words or phrasing to the original? You should choose the system that uses the most different words or word order.
\end{itemize}

\section{Reproducibility Notes}
\label{sec:reproducibility}

All experiments were run on a single Nvidia RTX 2080 Ti GPU. Training time for \textsc{Separator} was approximately 2 days on Paralex, and 1 day for QQP. \textsc{Separator} contains a total of 69,139,744 trainable parameters.

\section{Template Dropout}
\label{sec:templatedropout}

Early experiments showed that, while the model was able to separately encode meaning and form, the `syntactic' encoding space showed little ordering. That is, local regions of the encoding space did not necessarily encode templates that \textit{co-occurred} with each other in paraphrase clusters. We therefore propose \textit{template dropout}, where exemplars $\textbf{X}_{syn}$ are replaced with  probability $p_{td} = 0.3$ by a question with a different template from the same paraphrase cluster. This is intended to provide the model with a signal about which templates are similar to each other, and thus reduce the distance between their encodings.

\section{Ordering of the Encoding Space}
\label{sec:ordering}

\begin{figure}[!t]
    \centering
    \begin{subfigure}[b]{0.4\textwidth}
         \centering
         \includegraphics[width=\textwidth]{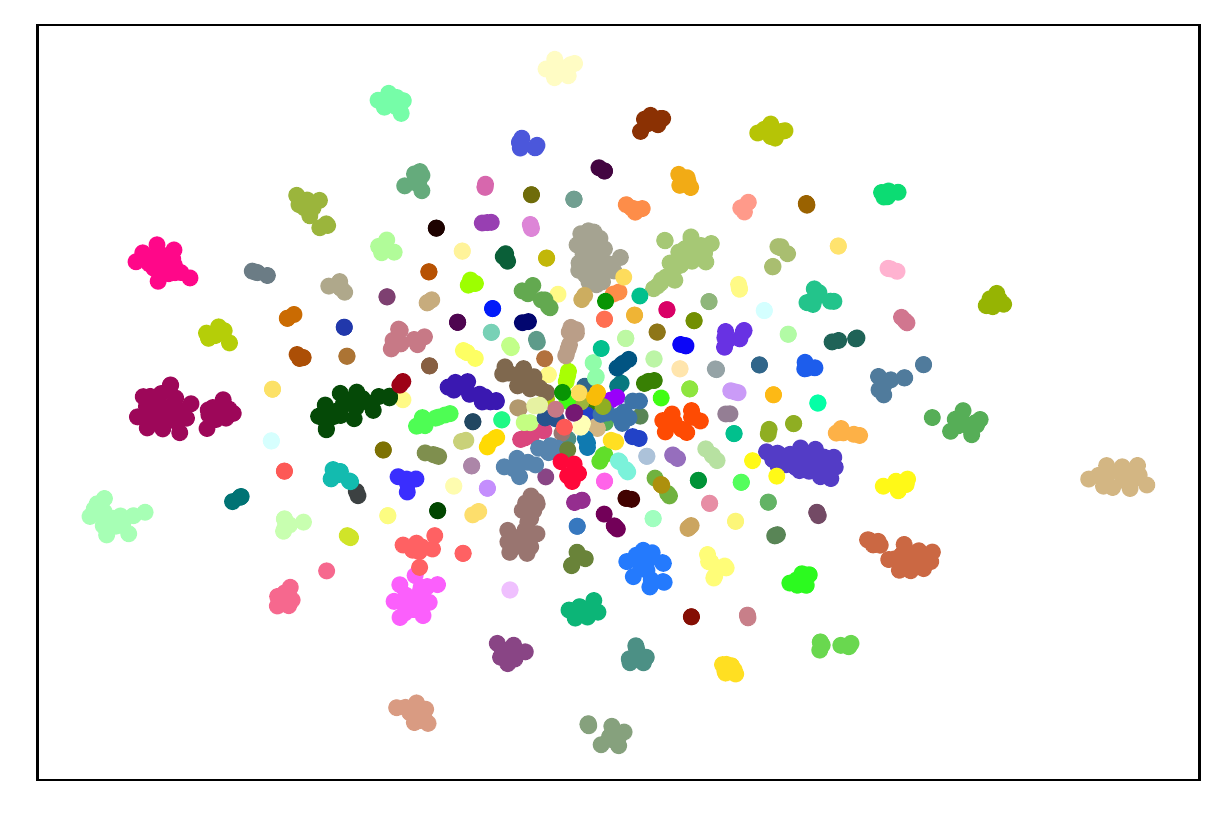}
         \caption{Semantic encodings}
         \label{fig:tsne_sem}
     \end{subfigure}
    \begin{subfigure}[b]{0.4\textwidth}
         \centering
         \includegraphics[width=\textwidth]{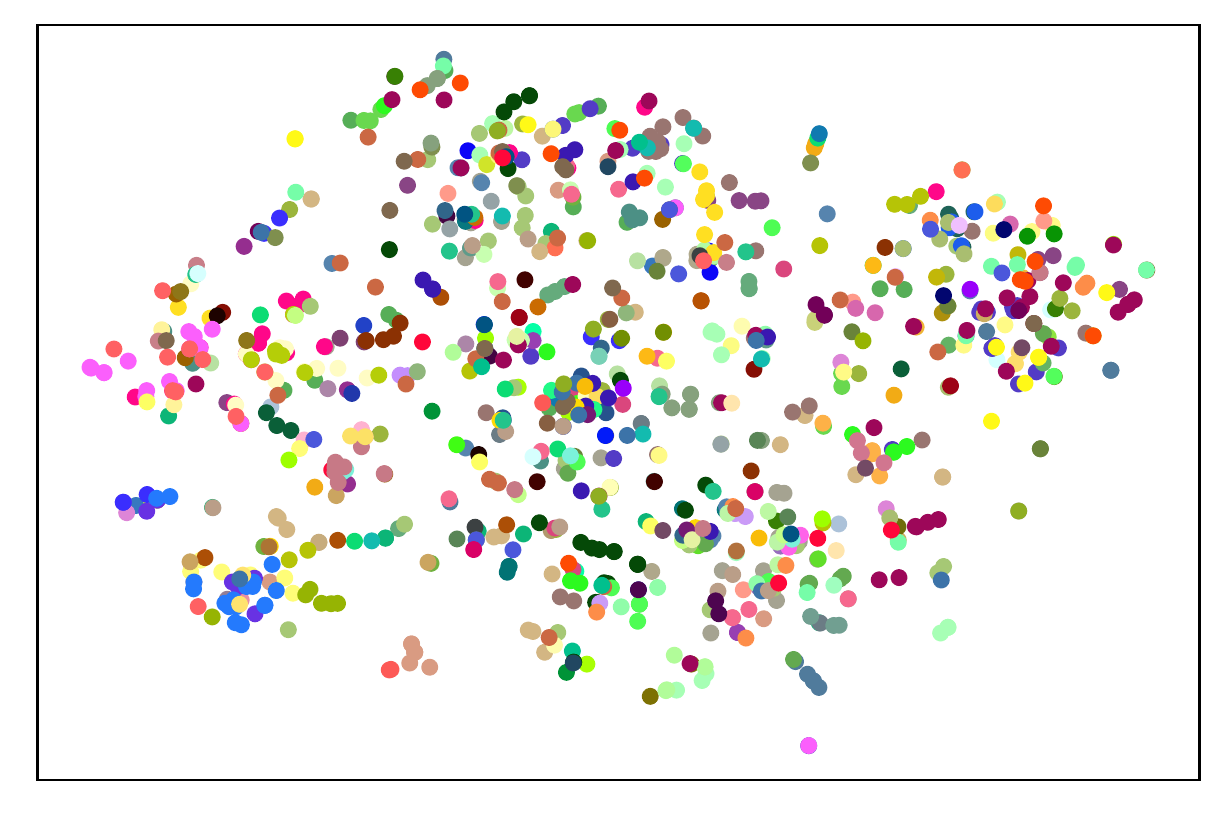}
         \caption{Syntactic encodings}
         \label{fig:tsne_syn}
     \end{subfigure}
    \caption{Visualisations of $z_{sem}$ and $z_{syn}$ using t-SNE \cite{tsne}, coloured by paraphrase cluster. The semantic encodings are clustered by meaning, as expected, but there is little to no local ordering in the syntactic space; valid surface forms of a particular question do not necessarily have  syntactic encodings near to each other.}
    \label{fig:tsne}
\end{figure}

\Cref{fig:tsne} shows that the semantic encodings $\textbf{z}_{sem}$ are tightly clustered by paraphrase, but the set of valid forms for each cluster overlaps significantly.  

In other words, regions of licensed templates for each input are not contiguous, and naively perturbing a syntactic encoding for an input question is not guaranteed to lead to a valid template. Template dropout, described in \Cref{sec:templatedropout}, seems to improve the arrangement of encoding space, but is not sufficient to allow us to `navigate' encoding space directly. The ability to induce an ordered encoding space and introduce syntactic diversity by simply perturbing the encoding, would allow us to drop the template prediction network, and we hope that future work will build on this idea.

\section{Failure Cases}
\label{sec:failure}

A downside of our approach is the use of an information bottleneck; the model must learn to compress a full question into a single, fixed-length vector. This can lead to loss of information or corruption, with the output occasionally repeating words or generating a number that is slightly different to the correct one, as shown in \Cref{tab:errors}.


We also occasionally observe instances of the well documented posterior collapse phenomenon, where the decoder ignores the input encoding and generates a generic high probability sequence.

\begin{table}[t!]
    \small
    \centering
    \begin{tabular}{lp{6cm}} \hline \hline
\multicolumn{2}{c}{\textbf{Numerical error}} \\
\emph{Input} & Replace starter on a 1988 Ford via? \\
\emph{Output} & How do you replace a starter on a 1992 Ford? \\
        \hline\hline
 \multicolumn{2}{c}{\textbf{Repetition}} \\
 \emph{Input} & What brought about the organization of the Republican
 political party? \\
\emph{Output} & What is the political party of the Republican party? \\
        \hline\hline
 \multicolumn{2}{c}{\textbf{Ignoring encoding}} \\
\emph{Input}& What do Hondurans do for a living?\\
\emph{Output}& What do Hondurans eat?\\ \hline \hline
    \end{tabular}
    \caption{Examples of failure modes.}
    \label{tab:errors}
\end{table}

\end{document}